\documentclass{article} % For LaTeX2e

\usepackage{times}
\usepackage{epsfig}

\usepackage{graphicx} % Allows including images
\usepackage{wrapfig}
\usepackage{cutwin}
\usepackage{animate}
\usepackage{booktabs} % Allows the use of \toprule, \midrule and \bottomrule in tables
\usepackage[T1]{fontenc}    % use 8-bit T1 fonts
\usepackage{hyperref}       % hyperlinks
\usepackage{url}            % simple URL typesetting
\usepackage{booktabs}       % professional-quality tables
\usepackage{amsfonts}       % blackboard math symbols
\usepackage{nicefrac}       % compact symbols for 1/2, etc.
\usepackage{microtype}      % microtypography
\usepackage{bm}
\usepackage{listings}
\usepackage{multirow}
\usepackage{caption}
\usepackage{subfig}
\usepackage{amsmath}
\usepackage{amsthm}
\usepackage{color}
\usepackage{xcolor}
\usepackage{colortbl}
\usepackage{amssymb}
\usepackage{mathtools}
\usepackage{pifont}
\usepackage{float}
\usepackage{verbatim}
\usepackage{lipsum}
\newcommand{\E}{\mathbb{E}}      
\newcommand{\vz}{{\bm z}}       
\newcommand{\sZ}{\mathcal{Z}}     
\newcommand{\vx}{{\bm x}}  
\newcommand{\vy}{{\bm y}}

\newcommand{\sR}{\mathbb{R}} 
\newcommand{\sS}{\mathbb{S}}

\newtheorem{thm}{Theorem}
\newtheorem{cly}{Corollary}

\usepackage[accepted]{icml2020}

\icmltitlerunning{Spherical Autoencoder}

\begin{document}

\twocolumn[
\icmltitle{Latent Variables on Spheres for Autoencoders in High Dimensions}

% It is OKAY to include author information, even for blind
% submissions: the style file will automatically remove it for you
% unless you've provided the [accepted] option to the icml2019
% package.

% List of affiliations: The first argument should be a (short)
% identifier you will use later to specify author affiliations
% Academic affiliations should list Department, University, City, Region, Country
% Industry affiliations should list Company, City, Region, Country

% You can specify symbols, otherwise they are numbered in order.
% Ideally, you should not use this facility. Affiliations will be numbered
% in order of appearance and this is the preferred way.
\icmlsetsymbol{equal}{*}

\begin{icmlauthorlist}
\icmlauthor{Deli Zhao}{equal,goo}
\icmlauthor{Jiapeng Zhu}{equal,to}
\icmlauthor{Bo Zhang}{goo}
\end{icmlauthorlist}

\icmlaffiliation{to}{Chinese University of Hong Kong}
\icmlaffiliation{goo}{Xiaomi AI Lab,}

\icmlcorrespondingauthor{Deli Zhao}{zhaodeli@gmail.com}
\icmlcorrespondingauthor{Jiapeng Zhu}{jengzhu0@gmail.com}
\icmlcorrespondingauthor{Bo Zhang}{zhangbo@xiaomi.com}

% You may provide any keywords that you
% find helpful for describing your paper; these are used to populate
% the "keywords" metadata in the PDF but will not be shown in the document
\icmlkeywords{Machine Learning, ICML}

\vskip 0.3in
]

% this must go after the closing bracket ] following \twocolumn[ ...

% This command actually creates the footnote in the first column
% listing the affiliations and the copyright notice.
% The command takes one argument, which is text to display at the start of the footnote.
% The \icmlEqualContribution command is standard text for equal contribution.
% Remove it (just {}) if you do not need this facility.

%\printAffiliationsAndNotice{}  % leave blank if no need to mention equal contribution
\printAffiliationsAndNotice{\icmlEqualContribution} % otherwise use the standard text.

\begin{abstract}
Variational Auto-Encoder (VAE) has been widely applied as a fundamental generative model in machine learning. For complex samples like imagery objects or scenes, however, VAE suffers from the dimensional dilemma between reconstruction precision that needs high-dimensional latent codes and probabilistic inference that favors a low-dimensional latent space.  By virtue of high-dimensional geometry, we propose a very simple algorithm, called Spherical Auto-Encoder (SAE), completely different from existing VAEs to address the issue. SAE is in essence the vanilla autoencoder with spherical normalization on the latent space. We analyze the unique characteristics of random variables on spheres in high dimensions and argue that random variables on spheres are agnostic to various prior distributions and data modes when the dimension is sufficiently high.  Therefore, SAE can harness a high-dimensional latent space to improve the inference precision of latent codes while maintain the property of stochastic sampling from priors. The experiments on sampling and inference validate our theoretical analysis and the superiority of SAE.
\end{abstract}

\section{Introduction}
Generative models, such as the autoencoder~\cite{Hinton06} and Variational Auto-Encoder (VAE)~\cite{VAE,VAE2}, play more and more important role for nonlinear dimensionality reduction and generation in machine learning and computer vision. The (variational) autoencoder has been a fundamental architecture of designing algorithms in deep learning. Our work will focus on the optimization of the autoencoder and make it more robust to prior distributions and the dimension of the latent space than VAE. 

Formally, suppose that $\bm x$ is a data point in the $d_x$-dimensional observable space $\mathbb{R}^{d_x}$ and $\bm y$  its  low-dimensional representation in the feature space $\mathbb{R}^{d_y}$, where $d_y \ll d_x$. The general formulation of the autoencoder can be written as \vspace{-0.1cm}
\begin{equation}\label{eq:AE}
 \text{Autoencoder:}~~\vx \overset{f}{\mapsto} \vy \overset{g}{\mapsto} \tilde{\vx},
\vspace{-0.1cm}
\end{equation} 
where $f(\cdot)$ and $g(\cdot)$ are the encoder and the decoder, respectively, and $\tilde{\vx}$ is the reconstruction of $\vx$. The $f(\vx)$ mapping can be viewed as nonlinear dimensionality reduction and the role of $g(\vy)$ as a regularizer to $f(\vx)$ in the autoencoder~\cite{Hinton06}.

VAE improves the vanilla autoencoder by posing a stochastic condition on the variables in  $\mathbb{R}^{d_y}$, such that the latent variables comply with a given prior distribution $\mathcal{P}$. According to convention, we let $\bm z$ represent the latent variable. Thus we can write the diagram of VAE as \vspace{-0.1cm}
\begin{equation}
  \text{VAE:}~~\vx \overset{f}{\mapsto} \vz \overset{g}{\mapsto} \tilde{\vx}, ~ \bm z  \sim \mathcal{P} ( \bm z). \vspace{-0.1cm}
\end{equation}
In the parlance of probability, the process of $\bm x  \mapsto \bm z = f(\bm x)$ is called inference, and the other procedure of $\bm z  \mapsto \tilde{\vx} = g(\bm z)$ is called sampling or generation. VAE is capable of carrying out one-pass inference and generation in one framework by two collaborative functional modules. An elegant algorithm was proposed by~\cite{VAE} to solve VAE via variational inference. However, a limitation of VAE is that it is sensitive to the dimension of the latent space and restrictive to the prior. We will give the analysis in section~\ref{sec:dimension}.

Using the geometric theory in this paper, we propose a simple method to improve the autoencoder with a latent space robust to stochastic sampling and dimension. Our theory is to reshape the latent space of the autoencoder on a sphere in high dimension, i.e. \vspace{-0.1cm}
\begin{equation}\label{eq:SAE1}
  \text{SAE:}~~\vx \overset{f}{\mapsto} \vz \overset{g}{\mapsto} \tilde{\vx}, ~ \vz \in \sS^{d_z-1}, ~ \vz^{\top} \bm 1=0, \vspace{-0.1cm}
\end{equation}
where $\sS^{d_z-1}$ is the sphere embedded  in $\mathbb{R}^{d_z}$ and $\bm 1$ is the all-one vector. Here we have no any probabilistic constraint on $\vz$.   With centerized $\vz$ on the sphere, we can rigorously prove that $\vz$ is robust to sampling on arbitrary prior distributions and varying dimensions. Our contributions are summarized as follows.  \vspace{-0.1cm}
\begin{itemize}
    \item The dimensional dilemma in VAE is analyzed when the dimension of the latent space is high.
    \item We introduce the volume concentration of high-dimensional spheres. Based on this property, we point out that projecting latent variable on a sphere is favorable of learning from the viewpoint of the volume in high-dimensional spaces.  \vspace{-0.1cm}
    \item We further introduce the probability distribution of distances between two arbitrary sets of random latent variables on the sphere in high dimensions and illustrate the phenomenon of distance convergence. Furthermore, we prove that the Wasserstein distance between two arbitrary sets of latent variables randomly drawn from a high-dimensional sphere are nearly identical, meaning that the variables on the sphere are distribution-robust. \vspace{-0.1cm}
    \item Based on our theoretical analysis, we propose a very simple algorithm, called Spherical Auto-Encoder (SAE), to improve the vanilla autoencoder.  The spherical normalization is simply put on latent variables instead of variational inference. In contrast to variational inference, we name the corresponding inference by SAE as spherical inference. 
\end{itemize} \vspace{-0.1cm}
We perform the experiments on MNIST letters and FFHQ faces to validate our theoretical analysis and claims with sampling and inference. 
\section{Dimensional Dilemma in VAE}\label{sec:dimension}
To be formal, we write the approximation of the marginal log-likelihood for VAE as\vspace{-0.1cm}
\begin{align}
  \log p(\vx)  & = \log \int p(\vx|\vz)p(\vz) d \vz \\
  &\geq -{\text{KL}}[q(\vz|\vx) || p(\vz)] + \E_q[\log p(\vx|\vz)]\label{eq:VAE},  \vspace{-0.2cm}  
\end{align}
where ${\text{KL}}[q(\vz|\vx) || p(\vz)]$ is the Kullback-Leibler divergence with respect to the posterior probability $q(\vz|\vx)$ and the prior $p(\vz)$. This lower bound is the objective to be optimized in VAE. The variational inference is an elegant solution to learn a stochastic latent space for an autoencoder. However, this probabilistic method suffers a critical limitation when the dimension of the latent space is high.

To understand this, we need to examine VAE from a dimensional view. The encoding operation $\vx \overset{f}{\mapsto} \vz$ can be regarded as the process of dimensionality reduction. To correctly reconstruct $\vx$ through the decoder, one condition is that the dimension $d_z$ of the latent space is no less than the intrinsic dimension of the underlying manifold where $\vx$ is drawn~\cite{Tenenbaum00,Roweis00,Bengio03}. Otherwise, the subspaces of the manifold will be folded after the encoder's projection and the reconstruction information will be lost, thus leading to impossibility of precise reconstruction via the decoder. To maintain the reconstruction precision, therefore, the autoencoder requires that $d_z$ should not be too low. From a probabilistic view, however, a large $d_z$ incurs the difficulty of fitting probabilistic distributions in high-dimensional latent spaces~\cite{Scott-1992,Handel-2016}. This phenomenon is called the curse of dimensionality that can be interpreted via a simple geometric fact. The volume ratio between a cube and its inscribed sphere goes to infinity when the dimension goes very large, meaning that the data points become rather sparse in high dimensions. Actually, the number of data points needed to fit a distribution grows exponentially when the dimension increases~\cite{Scott-1992}. Thus, the Kullback-Leibler divergence in (\ref{eq:VAE}) becomes challenging to measure the similarity between two distributions in high dimensions, provided a fixed number of data points. Therefore, $d_z$ is usually taken with much lower dimension compared with $d_x$ in VAE to void the curse of dimensionality.  This is the dimensional dilemma in VAE. Our work aims to solve this problem.

%\cite{KNN-KL-2007,treeKL-2013}

%
\begin{figure*}[htbp]
%\vskip 0.2in
\begin{center}
\begin{tabular}{ccc}
    \includegraphics[scale=0.7]{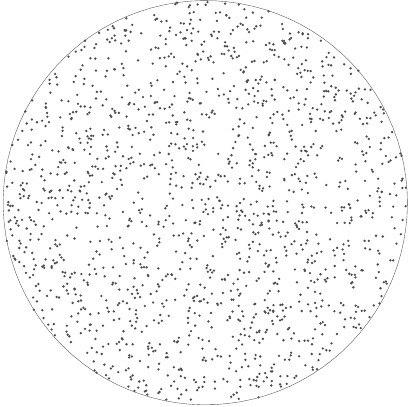}\hspace{0.cm}
    & 
    \includegraphics[scale=0.7]{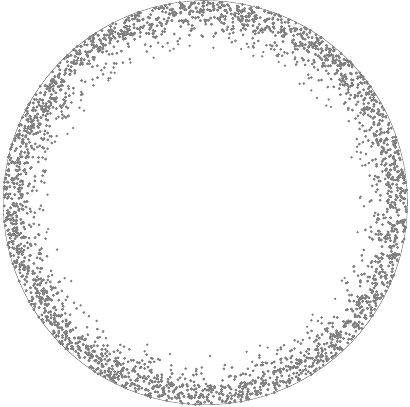}\vspace{0.cm}
    &
    \includegraphics[scale=0.65]{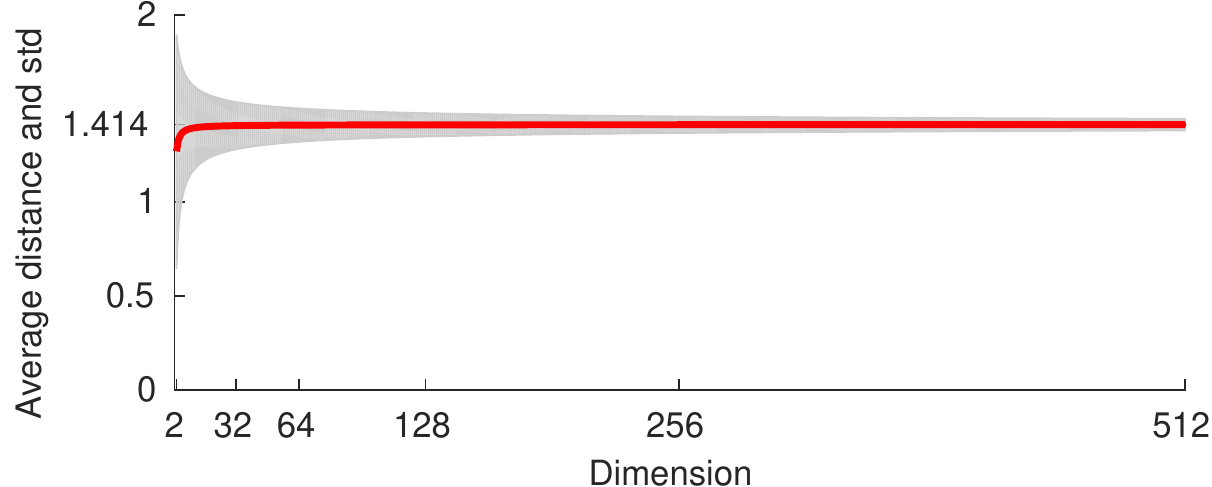}\\
    %
   % (a) Extractor training. & (b) GAN training. &  (c)  Encoder training.\vspace{0.cm}\\  
   (a) Low dimension.  &  (b) High dimension. & (c)  Distance convergence in high dimension. \vspace{0.cm}\\  
\end{tabular}
\end{center}\vspace{-0.3cm}
   \caption{ Geometry of spheres in high dimensions. (a) and (b) Volume of spheres in different dimensional spaces. The volume of the sphere in the high-dimensional space is highly concentrated near the surface. The interior is nearly empty. (c) Average distance between two points randomly sampled on unit spheres of various dimensions. The average distance is denoted by the red curve and the standard deviation by the gray background. The distances converge rapidly. They are nearly identical on the high-dimensional sphere.} \vspace{-0.3cm}
%\label{fig:volume}
\label{fig:geometry}
%\vskip -0.05in
\end{figure*}

\section{Latent Variables on Spheres}
For latent variables sampled from some priors, the projection on the unit sphere can can be easily performed by \vspace{-0.1cm}
\begin{equation}\label{eq:SN0}
%\vz \leftarrow \frac{\vz}{\| \vz \|}.
\vz \leftarrow \vz /\| \vz \|. \vspace{-0.1cm}
\end{equation}
This spherical normalization for priors fed into the generator is employed in StyleGAN that is the phenomenal algorithm in GANs~\cite{PGGAN,styleGAN18}. 
%To test the robustness of StyleGAN against the diverse distributions, we conduct two groups of experiments with the input $\vz$ sphere-normalized and not sphere-normalized when training StyleGAN. As shown in Figure~\ref{fig:variousz}, the diversity of generated faces is quite good for two different distributions with normalized $\vz$, whereas the face modes become similar for the case of the uniform distribution that $\vz$ is not normalized. 
In practice, we observe that StyleGAN with sphere-normalized $\vz$ is much more robust to the variation of variable modes from different distributions. Inspired by this observation, we interpret the benefit of using random variables on spheres by virtue of high-dimensional geometry in this section. Based on these geometric theories, a novel algorithm is proposed for improving autoencoder. 
\subsection{Volume Concentration}
For high-dimensional spaces, there are many counter-intuitive phenomena that will not happen in low-dimensional spaces. For a convenient analysis, we assume that the center of the sphere $\sS^d$ embedded in $\sR^{d+1}$ is at the origin. The concentration property of sphere volume is such intriguing counter-intuitive geometry.
%We first present the concentration property of sphere volume in $\sR^{d+1}$. 
%One can find the proof in~\cite{DataScience2020}. 
%
%\begin{thm}\label{thm:volume}
%Let $V(r)$ and $V\left((1-\epsilon)r\right)$ denote the volumes of the two concentric spheres of radius $r$ and $(1-\epsilon)r$, respectively, where $0<\epsilon<1$. Then\vspace{-0.1cm}
%
%\begin{equation}
%%    \frac{V\left((1-\epsilon)r\right)}{V(r)} = (1-\epsilon)^d. \vspace{-0.1cm}
%V\left((1-\epsilon)r\right)/V(r) = (1-\epsilon)^d. \vspace{-0.1cm}
%\end{equation}
%And if $\epsilon = t/d$, $ V\left((1-\epsilon)r\right)/V(r) \rightarrow e^{-t}$ when $d\rightarrow %\infty$, where $t$ is a constant.
%\end{thm}
%
Volume concentration says that the volume of the $d$-dimensional sphere of radius $(1-\epsilon)r$ ( $0<\epsilon<1$) rapidly goes to zero when $d$ goes large~\cite{DataScience2020}, meaning that the interior of the high-dimensional sphere is empty. In other words, nearly all the volume of the sphere is contained in the thin annulus of width $\epsilon r$. The width becomes very thin when $d$ grows. For example, the annulus of width that is $0.9\%$ of the radius contains $99\%$ of all the volume for the sphere in $\sS^{512}$. To help understand this counter-intuitive geometric property, we make a schematic illustration in Figures~\ref{fig:geometry} (a) and (b).

In fact, the  distributions defined on the sphere have been already exploited to re-formulate VAE, such as the von Mises-Fisher distribution~\cite{Tim-sphereVAE2018,Xu-sphereVAE2018}. But the algorithms proposed in~\cite{Tim-sphereVAE2018,Xu-sphereVAE2018} still fall into the category using the variational inference like the vanilla VAE, which also suffers from the dimensional dilemma. To eliminate this constraint, we need more geometric analysis.
\subsection{Distance Convergence} \label{sec:distance}
To dig deeper, we examine the pairwise distance between two arbitrary points randomly sampled on $\sS^{d}$. The following important theorem was proved by~\cite{Lord1954,sphereGame2002}.

\begin{thm}\label{thm:distance}
Let $\xi$ denote the Euclidean distance between two points randomly sampled on the sphere $\sS^{d}$ of radius $r$. Then the probability distribution of $\xi$ is \vspace{-0.2cm}
\begin{equation}
    \rho(\xi) = \frac{\xi^{d-2}}{c(d)r^{d-1}}\left[ 1- \left( \frac{\xi}{2r}  \right)^2  \right]^{\frac{d-3}{2}}, \vspace{-0.15cm}
\end{equation}
where the coefficient $c(d)$ is given by 
%\begin{equation}
   $ c(d) = \sqrt{\pi}\Gamma\left( \frac{d-1}{2}\right)/ \Gamma\left( \frac{d}{2}\right) $.
%\end{equation}
%
And the mean distance $\xi_{\mu}$ and the standard deviation $\xi_{\sigma}$ are  \vspace{-0.2cm}
\begin{equation}\label{eq:thm2}
  \xi_{\mu}  =\frac{ 2^{d-1} r \left[\Gamma\left( \frac{d}{2}\right) \right]^2 }   {  \sqrt{\pi} \Gamma\left( d-\frac{1}{2}\right) } \quad
  \text{  and  }   \quad
  \xi_{\sigma}  = \sqrt{2}r \sqrt{1-\frac{\xi^2_{\mu}}{2r^2}},\vspace{-0.2cm}
\end{equation}
respectively, where $\Gamma$ is the Gamma function. Furthermore, $\xi_{\mu} \rightarrow \sqrt{2}r \left( 1 - \frac{1}{8d} \right)$ and $\xi_{\sigma} \rightarrow \frac{r}{\sqrt{2d}}$ when $d$ goes large. \vspace{-0.2cm}
\end{thm}
Theorem~\ref{thm:distance} tells that the pairwise distances between two arbitrary points randomly sampled on $\sS^{d}$ approach to be mutually identical and converge to the mean $\xi_{\mu}=\sqrt{2}r$ when $d$ grows. The associated standard deviation $\xi_{\sigma} \rightarrow 0$.  We display the average distance and its standard deviation in Figure~\ref{fig:geometry}(c), showing that the convergence process is fast. Taking $\sS^{512}$ for example, we calculate that $\xi_{\mu}=1.4139$ and $\xi_{\sigma} = 0.0313$. The standard deviation is only $2.21\%$ of the average distance, meaning that the distance discrepancy between two arbitrary $\vz_i$ and $\vz_j$ on the sphere is rather small. This surprising phenomenon is also observed for neighborly polytopes when solving the sparse solution of underdetermined linear equation~\cite{Donoho-neighborly05} and for nearest neighbor search in high dimensions~\cite{NN99}.

With Theorem~\ref{thm:distance}, we can study the property of two different random datasets on $\sS^{d}$ pertaining to distribution-free sampling and spherical inference in generative models. 
Let $\sZ=\{\vz_1,\dots,\vz_n\}$ and $\sZ'=\{\vz'_1,\dots,\vz'_n\}$ be the datasets of random variables drawn from $\sS^{d}$ at \textit{random}, respectively. Our goal is to investigate the influence of two arbitrary different groups of latent variables on the autoencoder.  A rigorous way of quantifying the discrepancy between two datasets is the Wasserstein distance. 
To this end, we introduce the \textit{computational} definition of $2$-Wasserstein distance as\vspace{-0.1cm}
\begin{align}\label{eq:Wasserstein}
  W^2_2(\sZ,\sZ')  & = \min_{\bm \omega} \sum\nolimits^n_{i=1}\sum\nolimits^n_{j=1}\omega_{ij}\|\vz_i  -\vz'_j\|^2   \\
  & \text{s.t.} \quad {\textstyle \sum \nolimits}^n_{i=1}\omega_{ij} =  {\textstyle \sum \nolimits}^n_{j=1}\omega_{ij} = 1,\vspace{-0.2cm}
\end{align}
where $\bm \omega$ is the doubly stochastic matrix. Then we have
\begin{cly}\label{thm-wasserstein}
$W_2(\sZ,\sZ') \rightarrow \sqrt{2n} r$ with zero standard deviation when $d \rightarrow \infty$.
\end{cly}
Corollary~\ref{thm-wasserstein} is a direct result from Theorem~\ref{thm:distance} by substituting equation (\ref{eq:thm2}) into equation (\ref{eq:Wasserstein}).

Corollary~\ref{thm-wasserstein} says that despite the diverse data modes, the $2$-Wasserstein distance between two arbitrary sets of random variables randomly drawn on the sphere converges to a constant when the dimension is sufficiently large. For generative models, this unique characteristic  brings great convenience for distribution-robust sampling and spherical inference. For example, if $\sZ $ and $\sZ'$ obeys the different distributions, the functional role of $\sZ'$ nearly coincides with that of $\sZ$ with respect to Wasserstein distance, provided that both $\sZ $ and $\sZ'$ are randomly drawn from the high-dimensional sphere.  The specific distributions of $\sZ $ and $\sZ'$ affect the result negligibly under such a condition. We will present the specific application of Corollary~\ref{thm-wasserstein} in the following section.

In fact, we can obtain the bounds of $W_2(\sZ,\sZ')$ using the proven proposition about the nearly-orthogonal property of two random points on high-dimensional spheres~\cite{sphereAngle2013}. 
However, Corollary~\ref{thm-wasserstein} is sufficient to solve the problem raised in this paper. So, we bypass this analysis to simplify the theory for easy readability.

\subsection{Variable Centerization}\label{sec:centerization}
Both Theorem~\ref{thm:distance} and Corollary~\ref{thm-wasserstein} hold under one critical condition that latent vectors are \textit{randomly} drawn from the sphere. In practice, however,  this randomness for real data is hard to satisfy. For example, the condition violates if $\sZ $ is sampled from the open positive orthant and $\sZ'$ from the open negative orthant, or $\sZ $ from the normal distribution and $\sZ'$ from the Chi-squared distribution, etc. Hopefully, we can resort to central limit theorem to deal with it. For an arbitrary random vector $\vz_i$,  we let $\vz_i =[z^1_i,\dots,z^{d_z}_i]$ and the mean $\bar{z}_i = \frac{1}{d_z}\sum_j z^j_i$. Assume that $z^j_i$ is independent, identically distributed variables. Central limit theorem says that~\cite{Billingsley-1995}
\begin{equation}\label{eq:CLT}
%    \frac{\sum_j z^j_i - d_z \text{E}(z^j_i)}{\sqrt{d_z \text{var}(z^j_i)}} \sim \mathcal{N}(0,1)
     \sqrt{d_z}\left(\bar{z}_i - \text{E}(z^j_i)\right) /  \text{std}(z^j_i) \sim \mathcal{N}(0,1)
\end{equation}
when $d_z$ is sufficiently large. This conclusion is very meaningful for our case because the distribution of the mean can be the standard normal one despite the distribution of variable $z^j_i$. To satisfy the condition in Theorem~\ref{thm:distance} and Corollary~\ref{thm-wasserstein}, therefore, a very simple approach is that we only need to normalize latent variables by centerization $\vz_i - \bar{z}_i\bm 1$ and spherization $\vz_i /\| \vz_i \|$, on which is based our algorithm of spherical autoencoder that is  prior-agnostic.

\section{Spherical Autoencoder}
%
%We will present the SAE algorithm that is distribution-robust and prior-agnostic for spherical inference and sampling. 
%
According to Theorem~\ref{thm-wasserstein}, we may know that latent variables is agnostic to diverse distributions if they are randomly sampled from the high-dimensional sphere. Volume concentration guarantees that the error can be negligible even if they deviate from the sphere, as long as they are scattered near the spherical surface. This tolerance to various modes of latent random variables allow us to devise a simple solution to replace the variational inference for VAE. To be specific, we only need to constrain the centerized latent variables on the sphere by means of the standard framework of an autoencoder,  as opposed to the conventional way of employing the KL-divergence ${\text{KL}}[q(\vz|\vx) || p(\vz)]$ or its variants with diverse priors.  We can write the objective function for spherical autoencoder as \vspace{-0.1cm}
\begin{align}
    \min_{f,g}  &\| \bm x - \tilde{\vx}\|^2_{\ell_p}, \\ 
    \text{s.t.}  &  ~~\vz^{\top} \bm 1=0, ~ \vz \in \sS^{d_z-1}, \label{eq:SN} \vspace{-0.2cm}
\end{align}
where $\ell_p$ denotes the $p$-norm, $\vz = f(\vx)$, and $\tilde{\vx} = g(\vz)$. The constraint in (\ref{eq:SN}) can be fulfilled with spherical normalization, which is shown in the following the sequential mappings of the SAE framework\vspace{-0.1cm}
%
%\begin{equation}\label{eq:SAE}
%\underbrace{ \bm x \overset{f}{\longmapsto}  \bm z}_{\text{encoder}} \overset{}{\longmapsto} \underbrace{(\bm z - \hat{z}\bm 1) \overset{}{\longmapsto} \vz/{\|\vz\|}}_{\text{spherical normalization on the latent space}} \overset{}{\longmapsto} \underbrace{ \tilde{\vz} \overset{g}{\longmapsto} \tilde{\bm x} }_{\text{decoder}}, \vspace{-0.1cm}
%\end{equation}
%
%
\begin{equation}\label{eq:SAE}
\underbrace{ \bm x \overset{f}{\longmapsto}  \vz}_{\text{encoder}} \overset{}{\longmapsto} \underbrace{(\vz - \bar{z}\bm 1)/{\|\vz - \bar{z}\bm 1\|}}_{\text{spherical normalization}} \overset{}{\longmapsto} \underbrace{ \tilde{\vz} \overset{g}{\longmapsto} \tilde{\bm x} }_{\text{decoder}}, \vspace{-0.1cm}
\end{equation}
where $\bar{z} = \frac{1}{d_z}\sum_j z^j$ is the average of elements in $\vz$.
The objective function and the framework of our algorithm are much simpler than that of VAE and hyper-spherical VAE based on the variational inference or the variants of VAEs that apply various sophisticated regularizers on latent spaces. Our algorithm is purely geometric and free from the difficulty of any probability optimization. 

\section{Related work}
Little attention has been paid on examining geometry of latent spaces in the field of generative models. So we find few works directly related to ours. Most relevant one is S-VAE~\cite{Tim-sphereVAE2018,Xu-sphereVAE2018}, which applies the von Mises-Fisher (vMF) distribution as the probability prior. The vMF distribution is defined on the sphere.  The algorithms proposed in~\cite{Tim-sphereVAE2018,Xu-sphereVAE2018} both rely on the variational inference as VAE does. Therefore, S-VAE also suffers the dimensional dilemma and is restricted by specific priors.  

From the sampling viewpoint, our geometric analysis is directly inspired by ProGAN~\cite{PGGAN} and StyleGAN~\cite{styleGAN18} that have already applied spherical normalization (i.e. equation (\ref{eq:SN0})) for sampled inputs. We study the related theory and extend the case to devise a novel autoencoder that is free from the dimensional dilemma and is prior-agnostic. Another related  method is to sample priors along the great circle when performing the interpolation in the latent space for GANs~\cite{SampleGAN2016}.  This algorithm is perfectly compatible with our theory and algorithm. Therefore, it can also be harnessed in our algorithm when performing interpolation as well.

Wasserstein Auto-Encoder (WAE)~\cite{WAE2018} is an alternative way of optimizing the model distribution and the prior distribution using Wasserstein distance. Different from WAE,   SAE does not really use Wasserstein distance in the latent space. We just leverage Wasserstein distance to establish Corollary~\ref{thm-wasserstein} for the theoretical analysis. Adversarial Auto-Encoder (AAE)~\cite{AAE2015} is another interesting method of replacing the variational inference with adversarial learning in the latent space. But both WAE and AAE need some priors to match, which are essentially different from SAE.

Spherical Normalization (SN) in (\ref{eq:SAE}) is easily reminiscent of Batch Normalization (BN)~\cite{BN-2015} widely applied in deep learning. BN is performed among a batch of data points and there are learnable parameters, such that the normalization on data by BN relys on data modes or distributions. However, SN manipulates a single data point, independent of data distributions. Central limit theorem and Theorem~\ref{thm:distance} guarantee its plausibility. So SN and BN are established on different theory and for different purpose.

%
%\begin{equation}
%  q(\vz|\bm \mu,\kappa) = \mathcal{C}_{d_z}(\kappa) \exp({\kappa \bm \mu^{\top}\vz}),  \vspace{-0.1cm}  
%\end{equation}
%
%where $\kappa$ is the concentration coefficient for $\mu$ and $\mathcal{C}_{d_z}(\kappa)$ is the normalizing constant. 

%
%
\section{Experiment}\label{sec:experiment}
We conduct the experiments to test our theory and algorithms in this section. Three aspects pertaining to generative algorithms are taken into account, including sampling GANs, learning the variants of autoencoder, and sampling the decoders. 

The MNIST and  FFHQ datasets are used to evaluate algorithms. FFHQ~\cite{styleGAN18} is a more complex face dataset with large variations of faces captured in the wild.  We use the image size of $128\times 128$, which is larger than the commonly chosen size in the related work and also more challenging than $64\times 64$ or  $32\times 32$ for (variational) autoencoders to reconstruct. We test VAE and our SAE algorithm with this benchmark dataset for the case of high dimensions.
\subsection{Sampling GAN}\label{sec:sampling}
Our first experiment is to validate our theory and the robustness of our algorithm against diverse distributions for sampling. We employ StyleGAN trained with random variables sampled from the normal distribution. The other three different distributions are opted to test the generation with different priors after training, i.e. the uniform, Poisson, and Chi-squared distributions. The shapes of these three distributions are significantly distinctive from that of the normal distribution. Thus, the generalization capability of the generative models can be effectively unveiled when fed with priors that are not involved during training. We follow the experimental protocol in~\cite{PGGAN,styleGAN18} that StyleGAN is trained on the FFHQ face dataset and  Fr\'{e}chet inception distance (FID)~\cite{Borji18-metric} is used as the quality metrics of generative results. We take $d_z =512$, which is set in StyleGAN. This dimension is also used for both VAE and SAE on face data. 

From Table~\ref{tab:sampling}, we can see that the generative results by the normal distribution is significantly better than the others when tested with the original samples. The uniform distribution is as good as the normal distribution when projected on the sphere. This is because the values for each random vector are overall symmetrically distributed according to the origin. They satisfy the condition in Corollary~\ref{thm-wasserstein} after the spherical projection. The accuracy of Poisson and Chi-squared distributions is considerably improved after centerization, even better than the vanilla uniform distribution.  But the accuracy difference between all the compared distributions is rather negligible after centerization and spherization, empirically verifying the theory presented in Corollary~\ref{thm-wasserstein} and the distribution-agnostic property of our algorithm.
\begin{table}[t]
\centering
\caption{Comparison of sampling GAN on FFHQ face data. The quantitative results are FIDs. ``sph'' denotes spherization.}\label{tab:sampling} %\vspace{-0.cm}
  \begin{tabular}{  l  c  c  c  c }
    \hline
        & \multicolumn{2}{c}{no centerization} & \multicolumn{2}{c}{centerization}\\
    Distribution    & no sph & sph & no sph & sph \\
     \hline  %\vspace{0.1cm}
    Normal &   6.20   & 6.16  &  6.27 & \textbf{6.12}  \\
    Uniform &  33.93   &  6.16  &  33.86 &  \textbf{6.16} \\
    Poisson  &  23.70  &  26.85  & 18.15  & \textbf{6.19}  \\
    Chi &    25.26    &   27.07 & 11.34  & \textbf{ 6.16} \\
   \hline
  \end{tabular}\vspace{-0.cm}
\end{table}
\begin{figure*}[htbp]
%\vskip 0.2in
\begin{center}
\begin{tabular}{cccc}
    \includegraphics[scale=0.5]{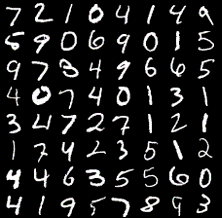}
    &
    \includegraphics[scale=0.5]{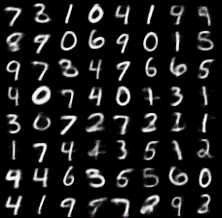}
    &
   \includegraphics[scale=0.5]{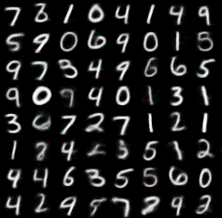} 
   &
   \includegraphics[scale=0.5]{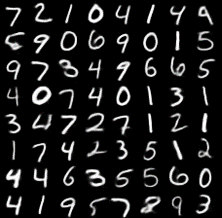}\\
   (a) MNIST samples & (b) VAE   &  (c) S-VAE   &  (d) SAE (ours)
   %\arrayrulecolor{blue}\hline
   \end{tabular}
\end{center}\vspace{-0.3cm}
   \caption{Reconstructed letters by VAEs and SAE with different priors on latent spaces.}\vspace{-0.3cm}
\label{fig:mnist-reconstruction}
\end{figure*}
\begin{figure*}[htbp]
\begin{center}
\begin{tabular}{ccc}
    \includegraphics[scale=0.43]{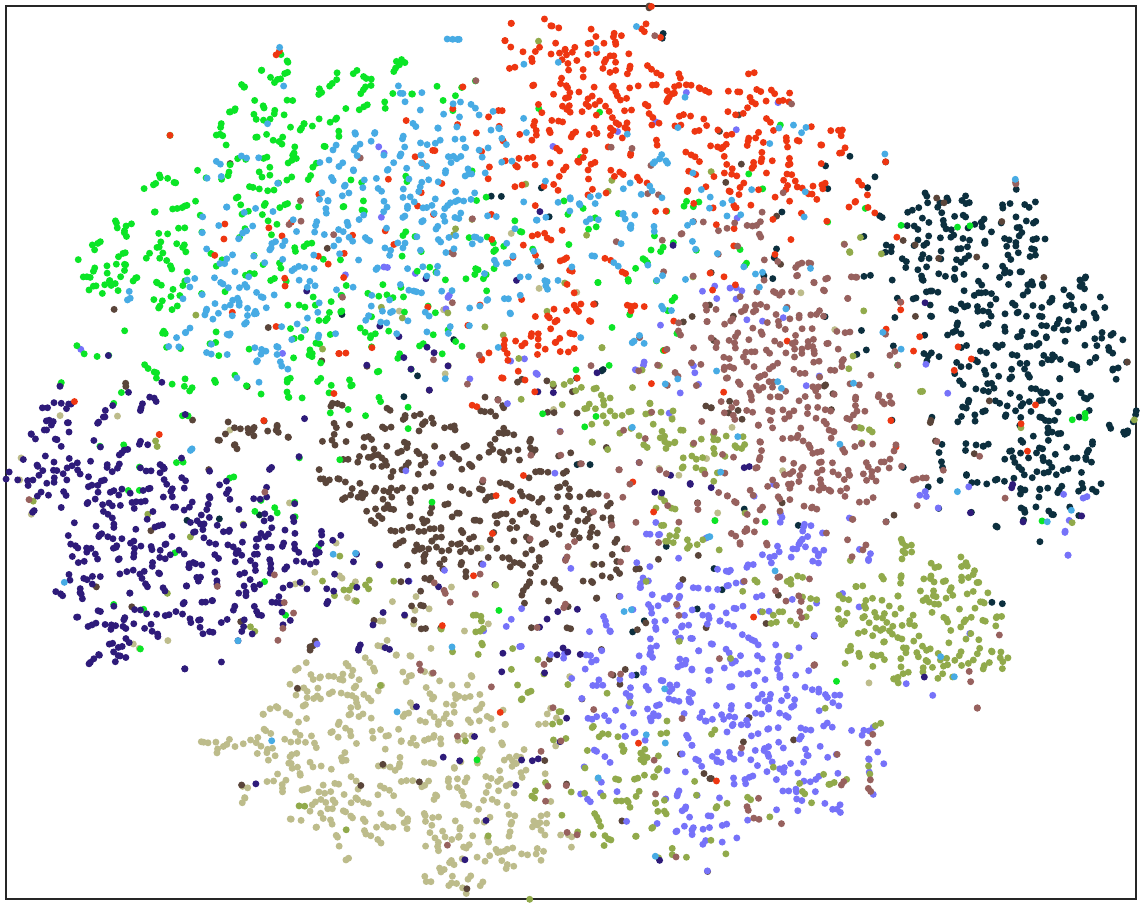}
    &
    \includegraphics[scale=0.43]{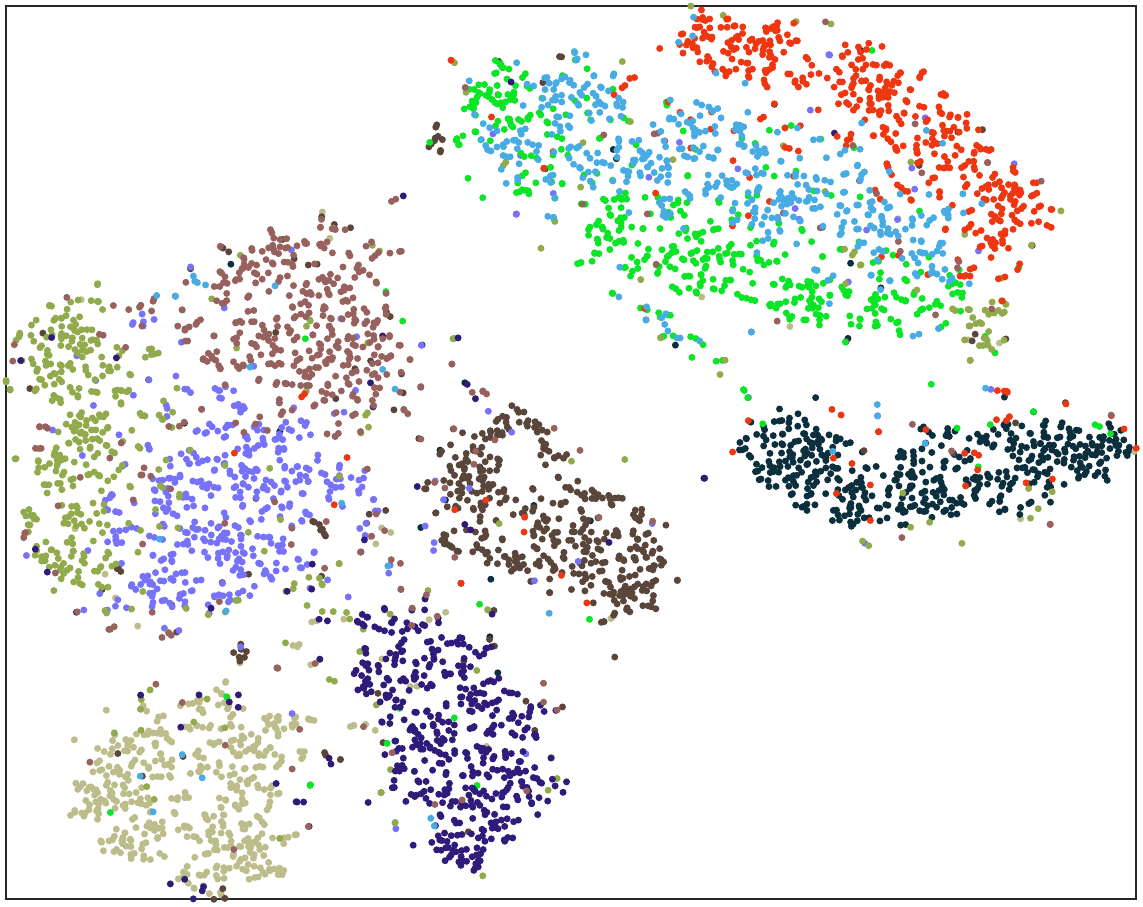}
     &
    \includegraphics[scale=0.43]{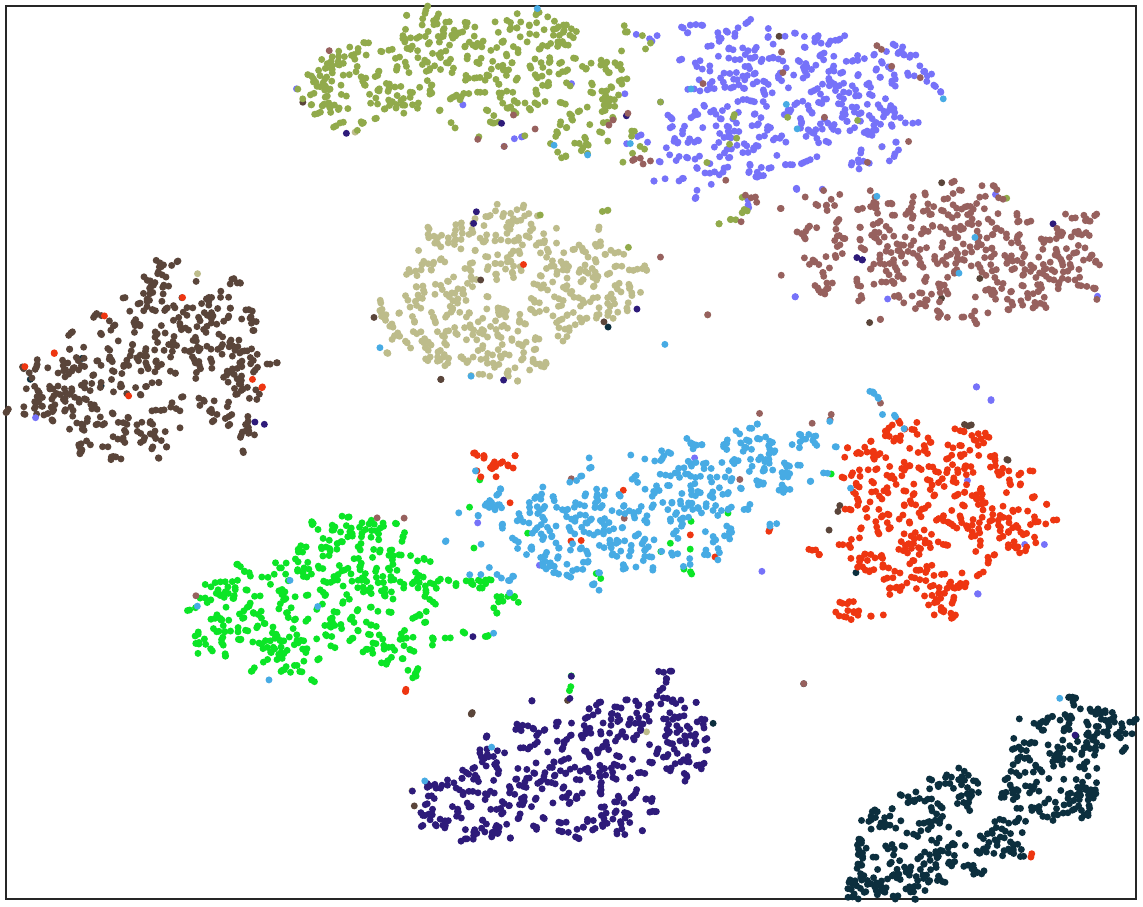}\\
    (a) VAE (normal) & (b) S-VAE (von Mises-Fisher)  & (c)  SAE (ours) \\
   \end{tabular}
\end{center}\vspace{-0.3cm}
   \caption{Visualization of inferred codes $\bm z$ on MNIST with t-SNE. We randomly sample 500 letters from each class in MNIST to form the whole set for illustration. }\vspace{-0.3cm}
\label{fig:visualization2}
\end{figure*}
\begin{figure*}[htbp]
%\vskip 0.2in
\begin{center}
\begin{tabular}{ccc}
    VAE  &  S-VAE  & SAE (ours)\\
    \includegraphics[scale=0.45]{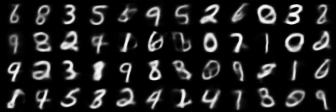}
    &
     \includegraphics[scale=0.45]{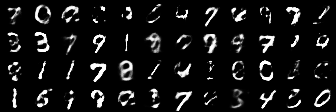}
    &
   \includegraphics[scale=0.45]{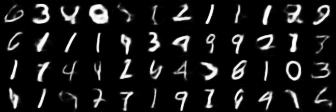}\\
    & Normal distribution & \\
     \includegraphics[scale=0.45]{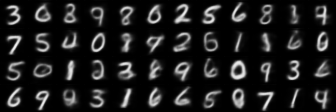}
    &
     \includegraphics[scale=0.45]{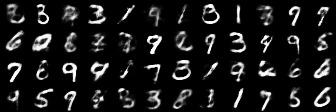}
    &
   \includegraphics[scale=0.45]{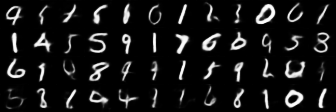}\\
    & Uniform distribution & \\
     \includegraphics[scale=0.45]{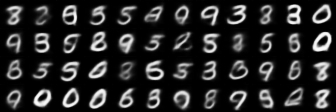}
    &
     \includegraphics[scale=0.45]{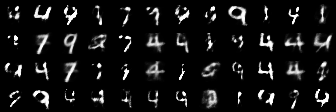}
    &
   \includegraphics[scale=0.45]{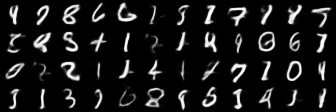}\\
    & Poisson distribution & \\
     \includegraphics[scale=0.45]{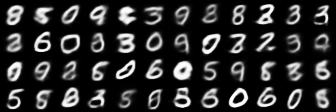}
    &
     \includegraphics[scale=0.45]{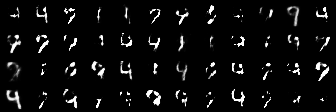}
    &
   \includegraphics[scale=0.45]{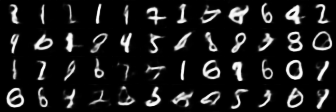}\\
    & Chi-squared distribution & \\
   \end{tabular}
\end{center}\vspace{-0.3cm}
   \caption{Generated letters with inputs of different priors. With the pre-trained decoders, the letters are generated with random vectors sampled from the four probability priors.  }\vspace{-0.3cm}
\label{fig:mnist-sampling}
\end{figure*}
\subsection{MNIST Letters}\vspace{-0.2cm}
We now use the commonly used MNIST dataset to learn the autoencoders of different styles, i.e. VAE, S-VAE, and our SAE. For all experiments on MNIST, we take $d_z =10$.

From Figure~\ref{fig:mnist-reconstruction}, we can see that the reconstruction letters by SAE are more faithful to the original ones than VAE and S-VAE.  For example, both VAE and S-VAE fail to recover the second letter ``2'' in the first row for each sub-figure while SAE obtains the accurate reconstruction.
To further reveal the advantage of SAE, we  visualize the latent codes of letters in  Figure~\ref{fig:visualization2} with t-SNE~\cite{tSNE}.  It is clear that the latent codes derived from SAE are much better than that from VAE and S-VAE. The margins between different classes are wider, meaning that the latent codes from the spherical inference conveys more discriminative information in the way of unsupervised learning. This experiment also indicates that SAE captures the intrinsic structure of multi-class data better than VAE and S-VAE. 
The superiority of SAE is more obvious when sampling the decoders after training, as Figure~\ref{fig:mnist-sampling} shows. For VAE, the sampling results from normal and uniform distributions are blurry and the mode aggregation occurs  for Poisson and Chi-squared distributions. For S-VAE, the sampling letters are worse than the reconstruction in Figure~\ref{fig:mnist-reconstruction}, implying that S-VAE is sensitive to priors. As a comparison, SAE performs consistently well on four priors, validating its robustness to different distributions and data modes. 
\begin{figure*}[htbp]
%\vskip 0.2in
\begin{center}
\begin{tabular}{cc}
     FFHQ faces & \includegraphics[scale=0.37]{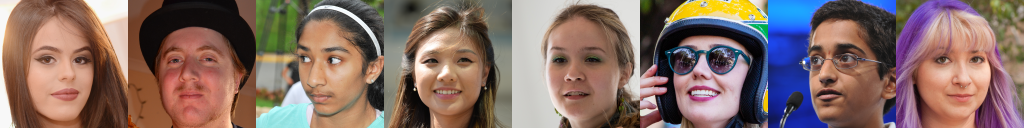}\\
    %  FFHQ faces \\
    %
    %LIA (ours) & \hspace{-0.5cm}
    VAE & \includegraphics[scale=0.37]{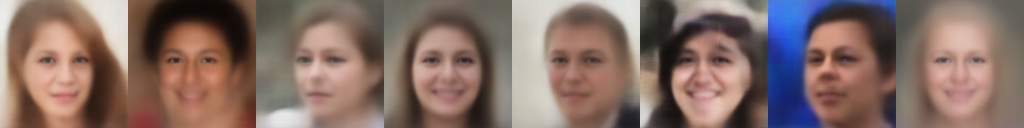}\\
    %VAE \\
    %ALI & \hspace{-0.5cm}
    %\includegraphics[scale=0.36]{figure/ffhq.png}\\
    %S-VAE (von Mises-Fisher)\\
    %
    SAE & \includegraphics[scale=0.37]{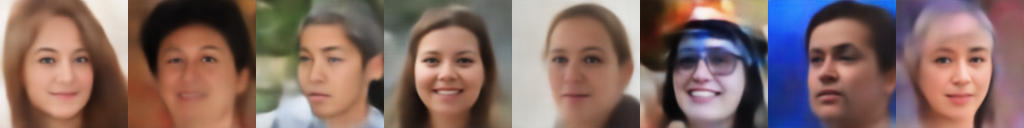}\\
    %SAE (ours) 
    %
   \end{tabular}
\end{center}\vspace{-0.3cm}
   \caption{Reconstructed faces by VAE and SAE. SAE only uses the spherical constraint in equation (\ref{eq:SAE}) instead of the variational inference in VAE. }\vspace{-0.3cm}
\label{fig:reconstruction}
\end{figure*}
\begin{figure*}[htbp]
%\vskip 0.2in
\begin{center}
\begin{tabular}{cc}
     VAE  &   SAE (ours) \\
    \includegraphics[scale=0.33]{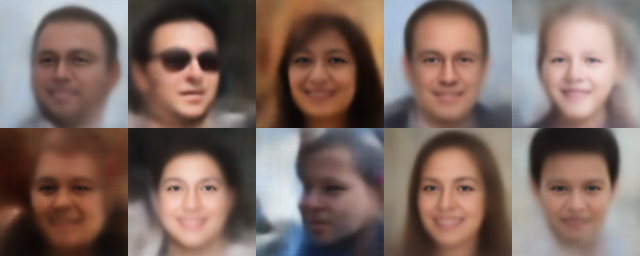} &
    %\hspace{-0.3cm}
    \includegraphics[scale=0.33]{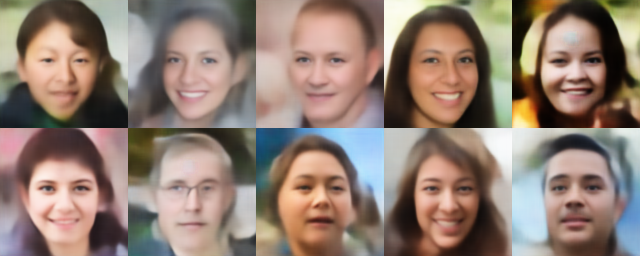}\\
     \multicolumn{2}{c}{Normal distribution}  \\
    \includegraphics[scale=0.33]{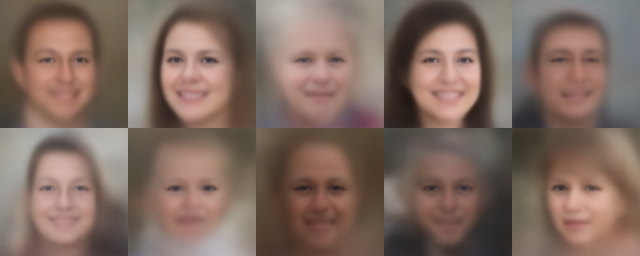} &
    \includegraphics[scale=0.33]{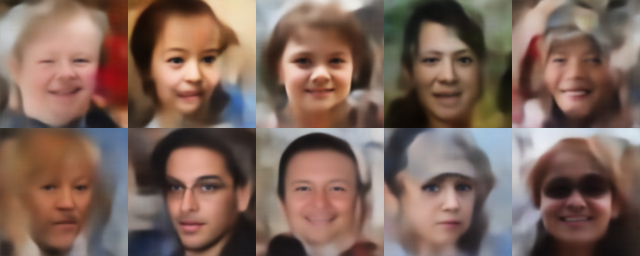}\\
    \multicolumn{2}{c}{Uniform distribution} \\
    \includegraphics[scale=0.33]{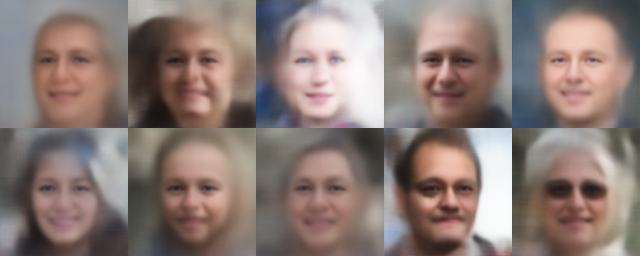} &
    \includegraphics[scale=0.33]{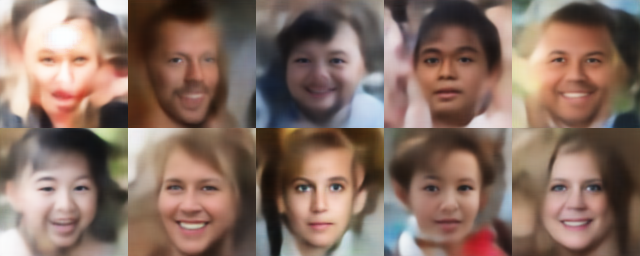}\\
   \multicolumn{2}{c}{Poisson distribution}\\
    \includegraphics[scale=0.33]{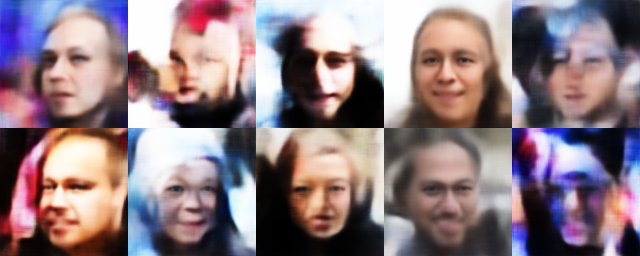} &
    \includegraphics[scale=0.33]{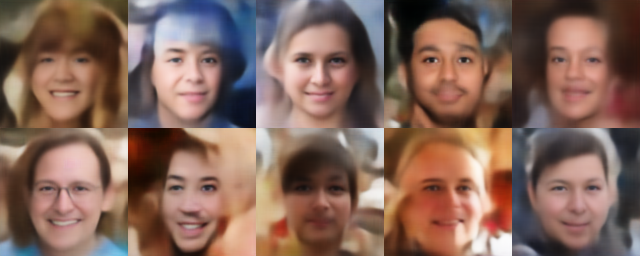}\\
    \multicolumn{2}{c}{ Chi-squared distribution }\\
   \end{tabular}
\end{center}\vspace{-0.3cm}
   \caption{Generated faces with inputs of different priors. With the pre-trained decoders of VAE and SAE, the faces are generated with the random vectors sampled from the four probability priors.  }\vspace{-0.3cm}
\label{fig:sampling2}
\end{figure*}
\begin{table}[htbp]
\centering
\caption{Quantitative comparison of face reconstruction.}\label{tab:accuracyConstruct} %\vspace{-0.1cm}
  \begin{tabular}{  l  c  c  c  }
    \hline
    Metric    & FID  & SWD & MSE\\
     \hline %\vspace{0.1cm}
     VAE &  134.22  & 77.68   & 0.091 \\
    SAE (ours) & 91.02  &  56.58  &  0.063    \\
     \hline
  \end{tabular}\vspace{-0.3cm}
\end{table}
\begin{figure*}[t]
%\vskip 0.2in
\begin{center}
\begin{tabular}{cc}
   VAE  \includegraphics[scale=0.37]{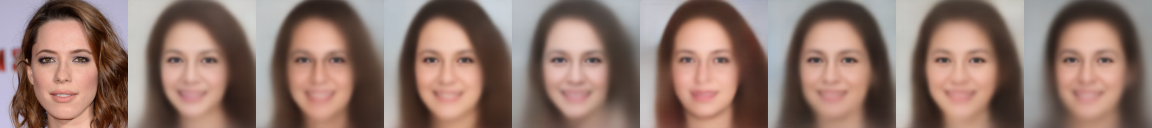}\\
   SAE  \includegraphics[scale=0.37]{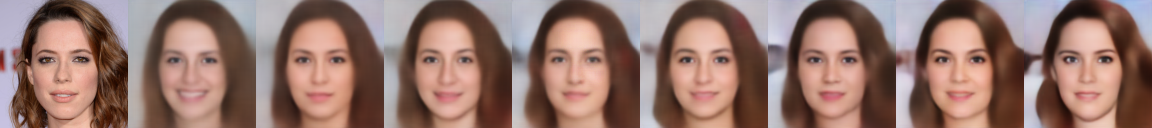}\\
   \end{tabular}
\end{center}\vspace{-0.3cm}
   \caption{Reconstructed faces associated with latent dimensions in Figure~\ref{fig:dimension}.}\vspace{-0.3cm}
\label{fig:dimension-face}
\end{figure*}
\begin{figure}[t]
%\vskip 0.2in
\begin{center}
     \includegraphics[scale=0.9]{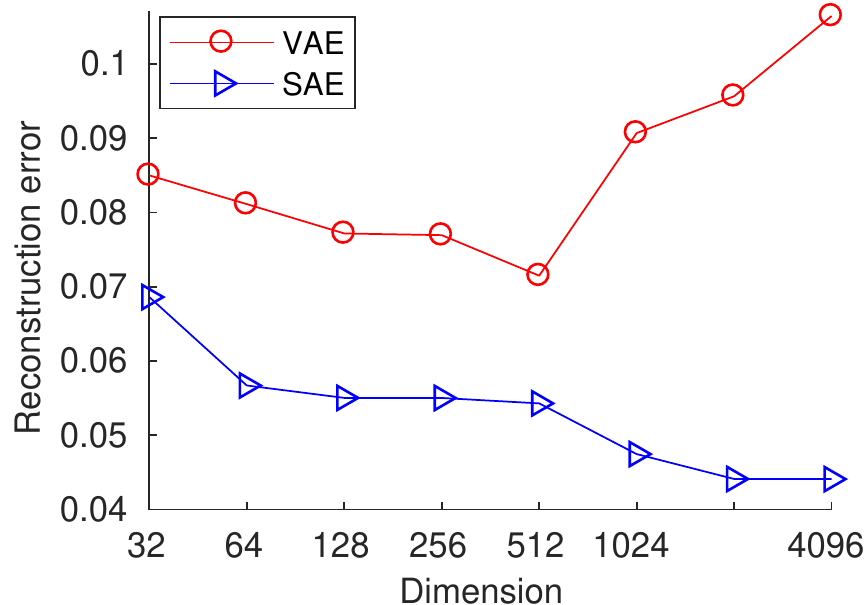}
\end{center}\vspace{-0.3cm}
   \caption{Reconstruction error as the function of dimensions.}\vspace{-0.3cm}
\label{fig:dimension}
\end{figure}
\subsection{FFHQ faces}
We compare the vanilla VAE with the normal prior~\cite{VAE} with our SAE algorithm for reconstruction and sampling tasks on face data in this section. S-VAE is not compared because we fail to train S-VAE on the FFHQ dataset
%
%
%\subsubsection{Reconstruction}
%
From Figure~\ref{fig:reconstruction}, we can see that the face quality of SAE  outperforms that of VAE. The imagery details like semantic structures are preserved much better for SAE. 
For example, the sunglasses in the sixth image is successfully recovered by SAE, whereas VAE distorts the face due to this occlusion.
It is worth emphasizing that the blurriness for images reconstructed by SAE is much less than that by VAE, implying that the spherical inference is superior to the variational inference in VAE. The different accuracy measurements in Table~\ref{tab:accuracyConstruct} also indicate the consistently better performance of SAE.
%

%
%\subsubsection{Sampling}
%
%
To test the generative capability of the models, we also perform the experiment of sampling the decoders as done in section~\ref{sec:sampling}. Prior samples are drawn from the normal, uniform, Poisson, and Chi-squared distributions, respectively, and then fed into the decoders to generate faces. Figure~\ref{fig:sampling2} illustrates the generated faces of significantly different quality with respect to four types of samplings. The style of the generated faces by SAE keeps consistent, meaning that SAE is rather robust to different probability priors. This also empirically verifies the correctness of Theorem~\ref{thm-wasserstein} by solving the real problem. As a comparison, the quality of the generated faces by VAE varies with probability priors. In other words, VAE is sensitive to the outputs of the encoder with the variational inference, which is probably the underlying reason of the difficulty of training VAE with sophisticated architectures.  We also present the experimental results on CelebA in supplementary material.
%
%
\iffalse
%
\begin{figure*}[t]
%\vskip 0.2in
\begin{center}
\begin{tabular}{cc}
    %
     \includegraphics[scale=0.8]{cdimension-face.pdf}
     & \includegraphics[scale=0.8]{cdimension-face.pdf}\\
   (a) MNIST  &  (b) FFHQ\\
   \multicolumn{2}{}{
   VAE ~~ \includegraphics[scale=0.4]{original_0.png}
   }\\
   \multicolumn{2}{}{
   SAE ~~ \includegraphics[scale=0.4]{original_0.png}
   }
    %
   \end{tabular}
\end{center}\vspace{-0.3cm}
   \caption{Reconstruction error as the function of dimensions.}\vspace{-0.3cm}
\label{fig:dimension}
\end{figure*}
%
\fi
%
%
\subsection{Varying Latent Dimensions}
To investigate the effectiveness of SAE to circumvent the dimensional dilemma, we analyze the results of varying the dimension of the latent spaces for VAE and SAE. As shown in Figure~\ref{fig:dimension}, SAE is capable of monotonically  decreasing the reconstruction error when the latent dimension grows. As a comparison, VAE's reconstruction error begins to increase when the dimension is larger than 512, meaning that the curse of dimensionality occurs. 
For VAE, the latent codes of faces beyond 512 dimensions  are too high to be applicable for variational inference. It is also obvious that the high latent codes of SAE produce significantly better performance than that of VAE, implying that the potential capability of the autoencoder can be unlocked if the curse of dimensionality posed on the latent \textit{random} space can be eliminated.
Figure~\ref{fig:dimension-face}  illustrates that the reconstructed faces by SAE are consistently better than that by VAE. More examples are attached in supplementary material.

%It is worth emphasizing that if the latent dimension grows, the reconstruction precision can also be improved for the vanilla autoencoder in (\ref{eq:AE}). But the vanilla autoencoder does not admit stochasticity for its latent space as VAE and SAE do.

%
\section{Conclusion}\vspace{-0.1cm}
In this paper, we attempt to address the curse of dimensionality for VAE. By analyzing the geometry of volume concentration and distance convergence on the high-dimensional sphere, we prove that the Wasserstein distance converges to be a constant for two datasets randomly sampled from the sphere when the dimension goes large. These unique characteristics imply that two random datasets drawn on the high-dimensional sphere are distribution-agnostic.  Based on this theory, 
we propose a very simple algorithm called Spherical Auto-Encoder (SAE). SAE is a standard autoencoder with spherical normalization on the latent space. 
The experiments on the MNIST letter and  FFHQ face databases validate the effectiveness of our theory and new algorithm for sampling and spherical inference. 

It is worth noting that the applications of our theory and algorithm are not limited for autoencoders. Interested readers may explore the possibility in other scenarios.

%For example, the extracted features by convolutional neural networks (CNN) are normalized prior to classification and clustering for face recognition~\cite{schroff2015facenet}. The underlying connection may exist with softmax optimization in such scenarios. Moreover, batch normalization (BN) is a basic tool in neural networks to  reduce internal covariate shift in data~\cite{BN:2015}. The effect of BN is in some extent relevant to centerization and spherization. Therefore, it is intriguing to investigate the principle of BN with spherical geometry.

%\subsubsection*{Acknowledgments}

%Use unnumbered third level headings for the acknowledgments. All
%acknowledgments, including those to funding agencies, go at the end of the %paper.

\bibliography{sphere}
\bibliographystyle{icml2020}

\clearpage
\appendix
%\section{Appendix}
\onecolumn
\section{Reconstruction on FFHQ}%\vspace{-3cm}
\begin{figure*}[htbp]
%\vskip 0.2in
\begin{center}
    \includegraphics[scale=0.64]{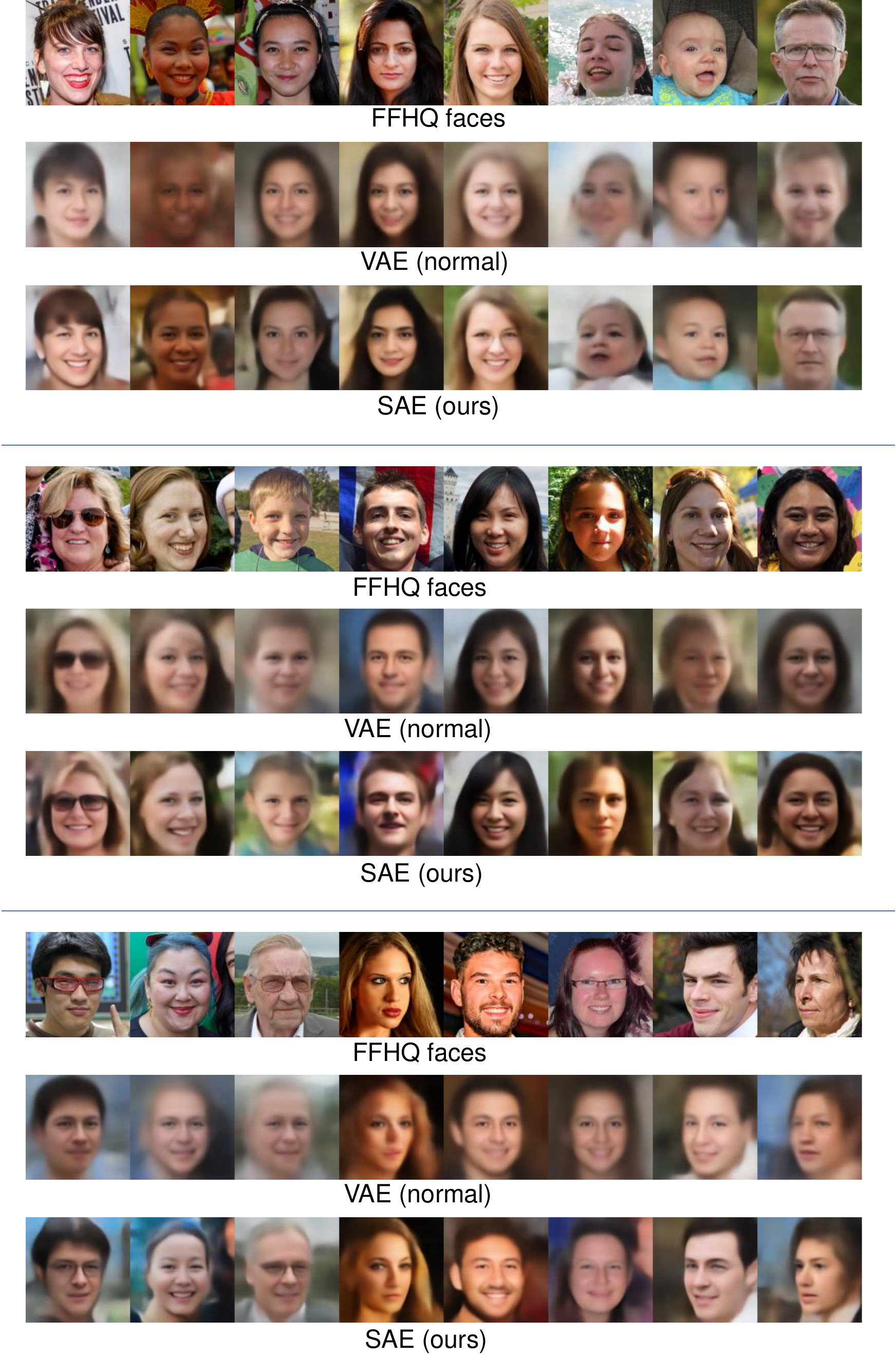}\\
\end{center}\vspace{-0.3cm}
   \caption{Reconstructed faces by VAE and SAE.}%\vspace{-0.3cm}
\label{fig:reconstruction4}
\end{figure*}

\clearpage
\onecolumn
\section{Sampling VAE and SAE on FFHQ}

\begin{figure*}[htbp]
%\vskip 0.2in
\begin{center}
\begin{tabular}{cc}
    \includegraphics[scale=0.75]{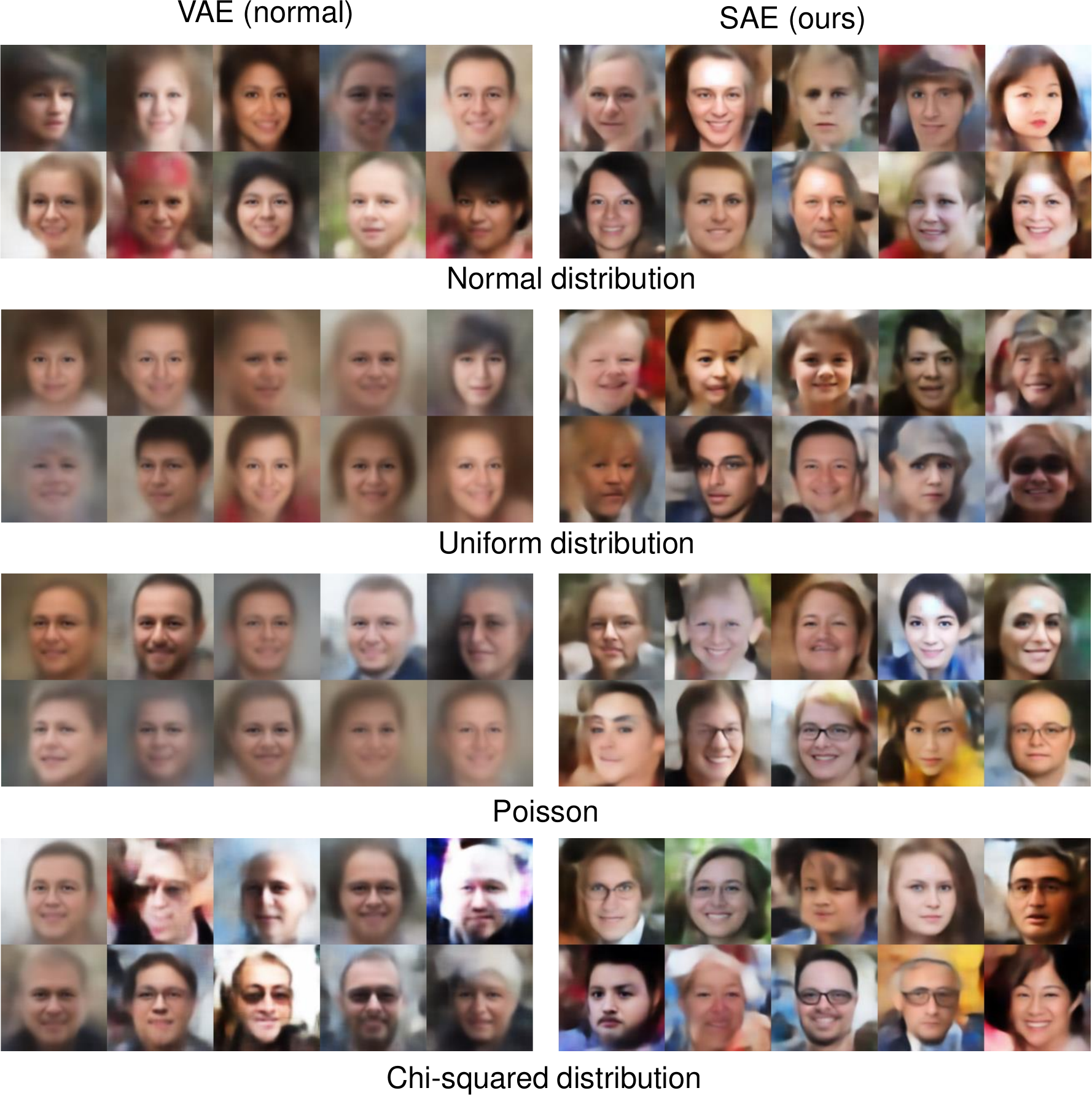}\\
   \end{tabular}
\end{center}\vspace{-0.3cm}
   \caption{Generated faces with inputs of different priors. With the pre-trained decoders of VAE and SAE, the faces are generated with random vectors sampled from the four probability priors.  }\vspace{-0.3cm}
\label{fig:sampling3}
\end{figure*}
\begin{figure*}[htbp]
%\vskip 0.2in
\begin{center}
\begin{tabular}{cc}
    \includegraphics[scale=0.75]{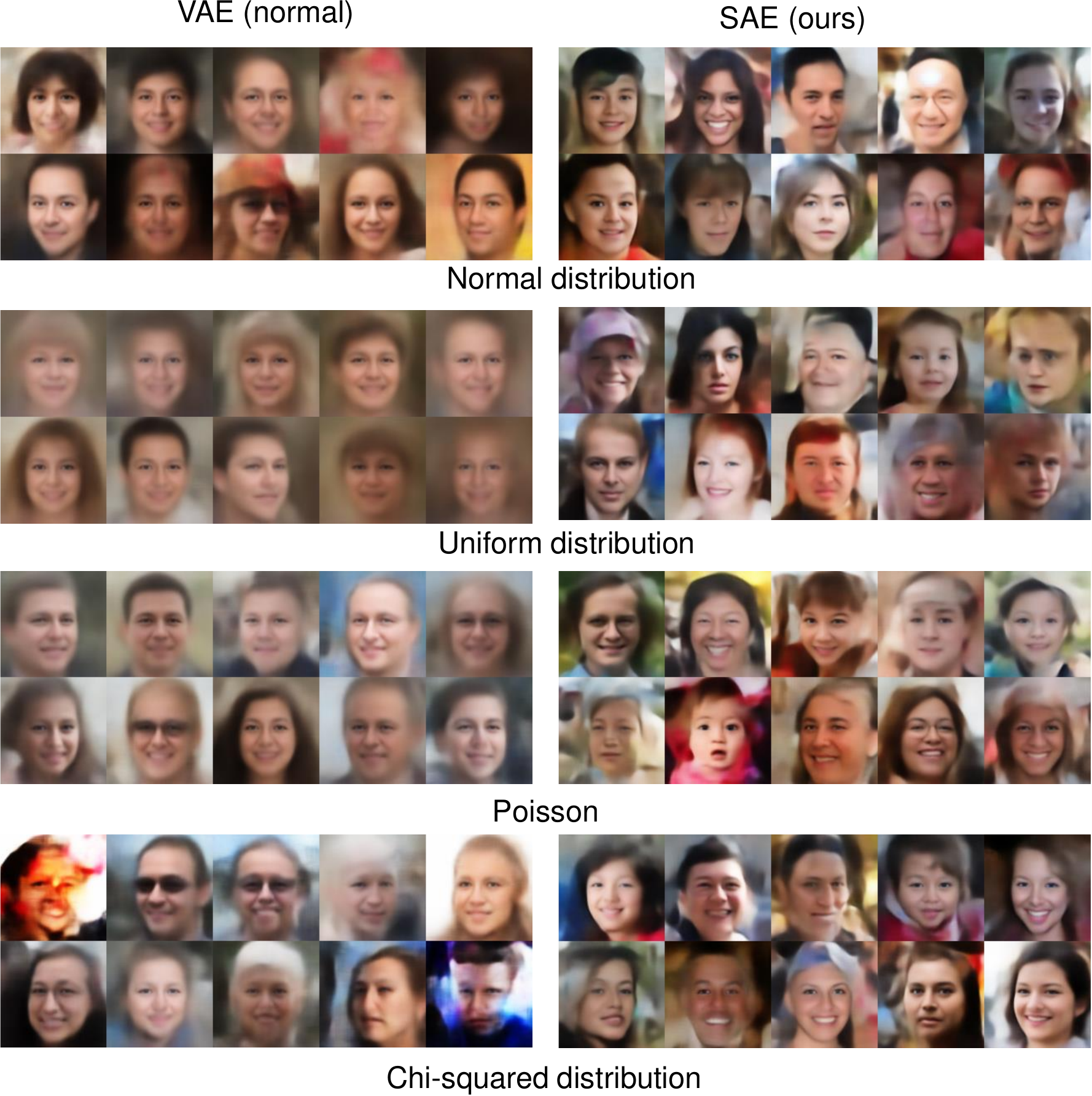} \\
   \end{tabular}
\end{center}\vspace{-0.3cm}
   \caption{Generated faces with inputs of different priors. With the pre-trained decoders, the faces are generated with random vectors sampled from the four probability priors. }\vspace{-0.3cm}
\label{fig:sampling4}
\end{figure*}
\clearpage
\onecolumn
\section{Reconstruction on CelebA}
\begin{figure*}[htbp]
%\vskip 0.2in
\begin{center}
\begin{tabular}{c}
    \includegraphics[scale=0.68]{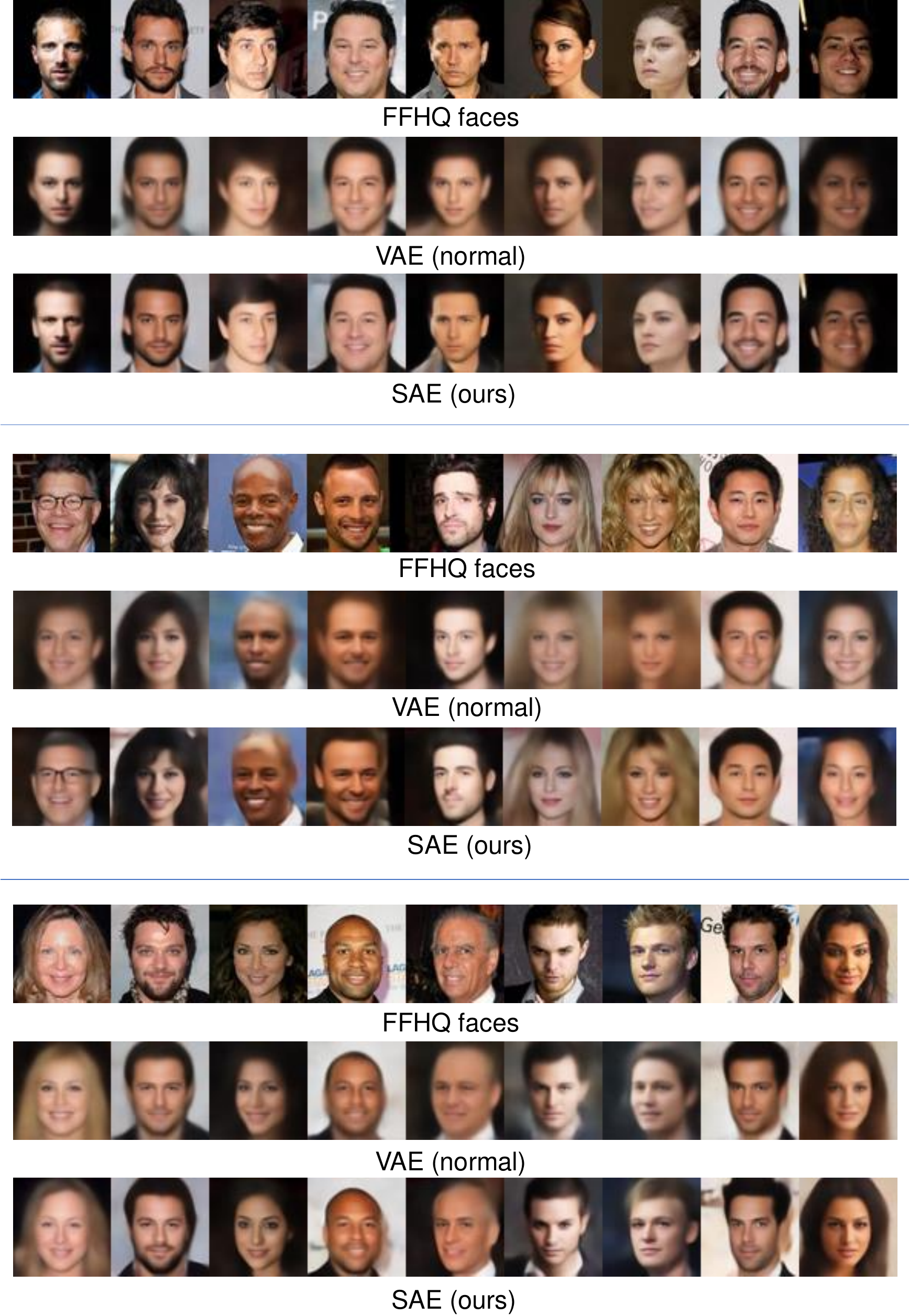}\\
   \end{tabular}
\end{center}\vspace{-0.3cm}
   \caption{Reconstructed faces by VAE and SAE.}\vspace{-0.3cm}
\label{fig:reconstruction3}
\end{figure*}

\clearpage
\onecolumn
\section{Sampling on CelebA}
\begin{figure*}[htbp]
%\vskip 0.2in
\begin{center}
\begin{tabular}{c}
    \includegraphics[scale=0.7]{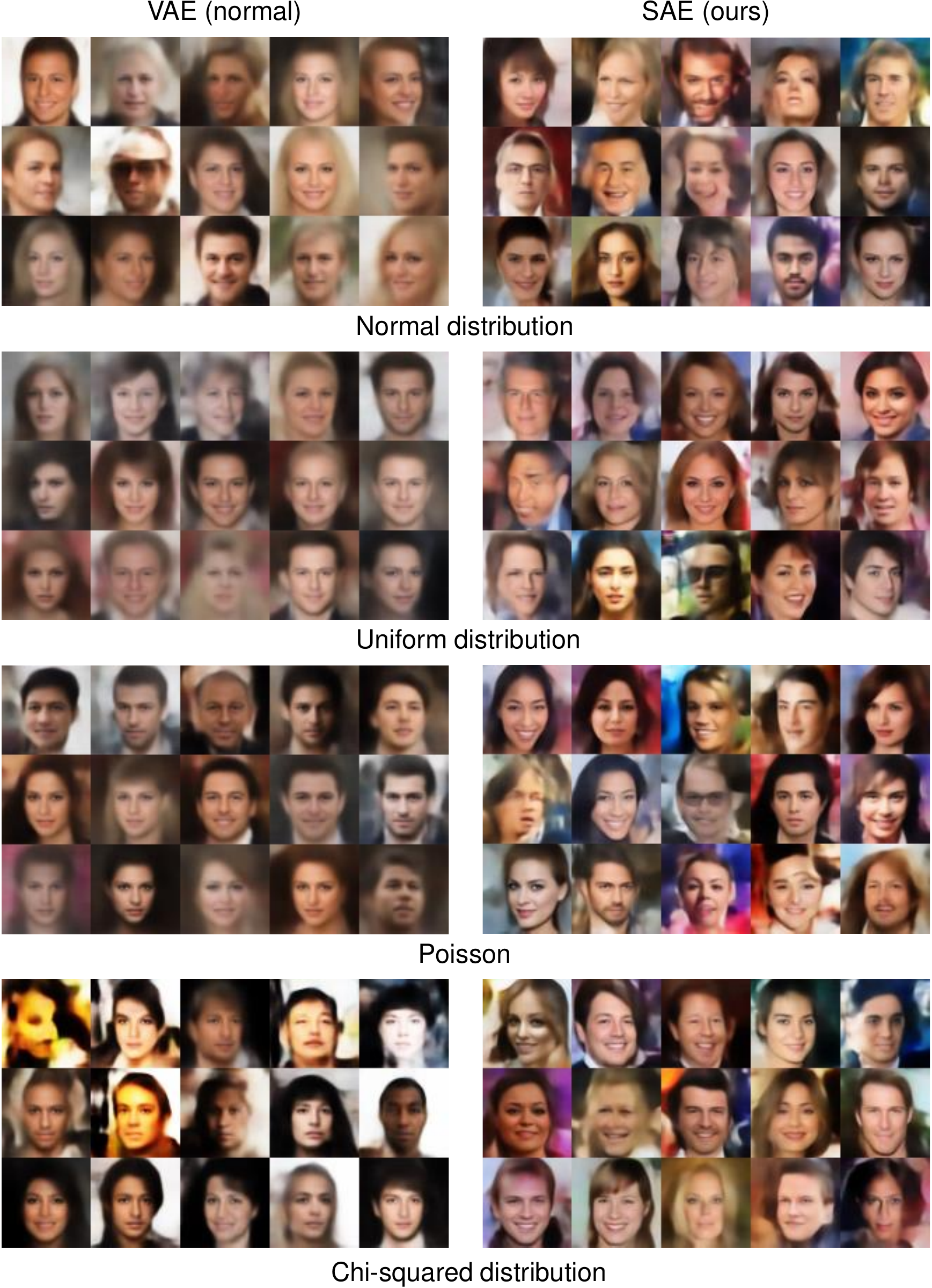} \\
   \end{tabular}
\end{center}\vspace{-0.3cm}
   \caption{Generated faces with inputs of different priors. With the pre-trained decoders, the faces are generated with random vectors sampled from the four probability priors. }\vspace{-0.3cm}
\label{fig:sampling6}
\end{figure*}

\clearpage
\onecolumn
\section{Visualization of Inference}
\begin{figure*}[htbp]
%\vskip 0.2in
\begin{center}
\begin{tabular}{c}
    \includegraphics[scale=0.7]{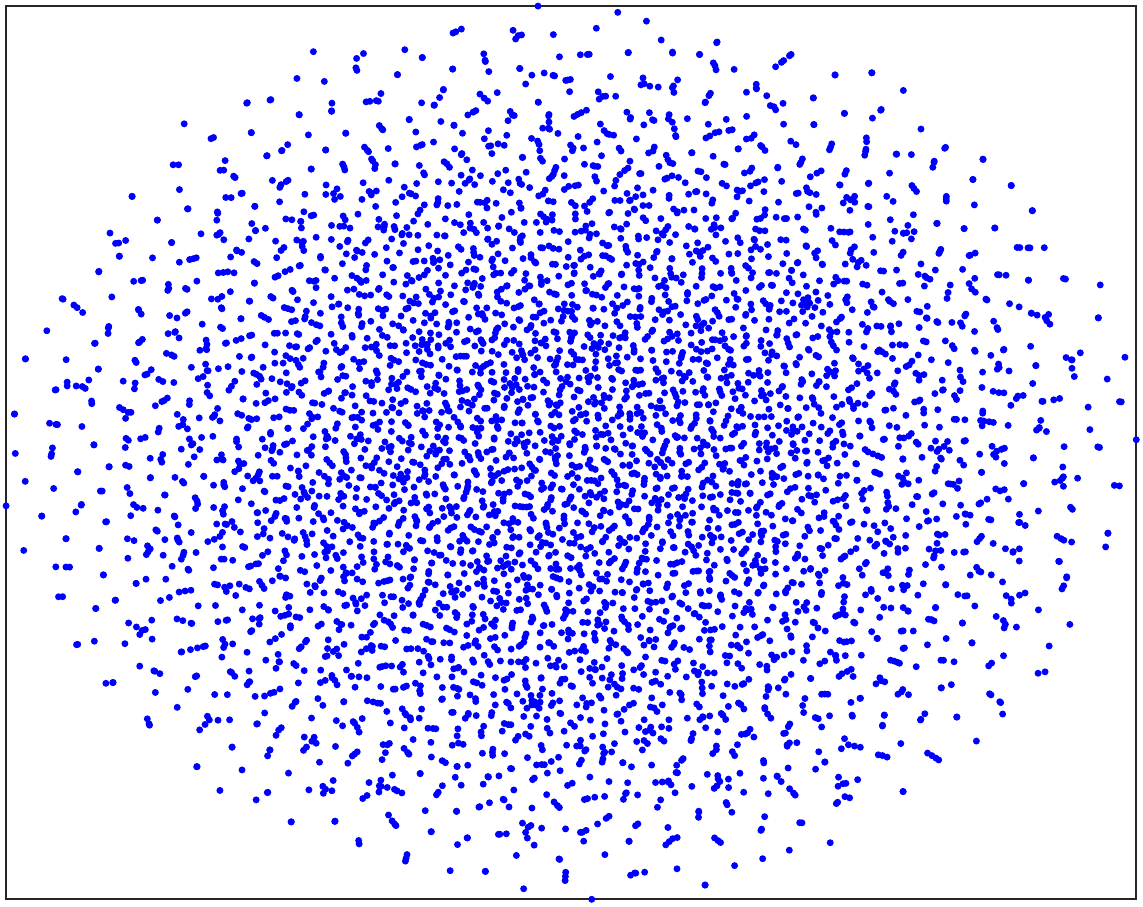}\\
     (a) VAE \\
    %
    %LIA (ours) & \hspace{-0.5cm}
    \includegraphics[scale=0.7]{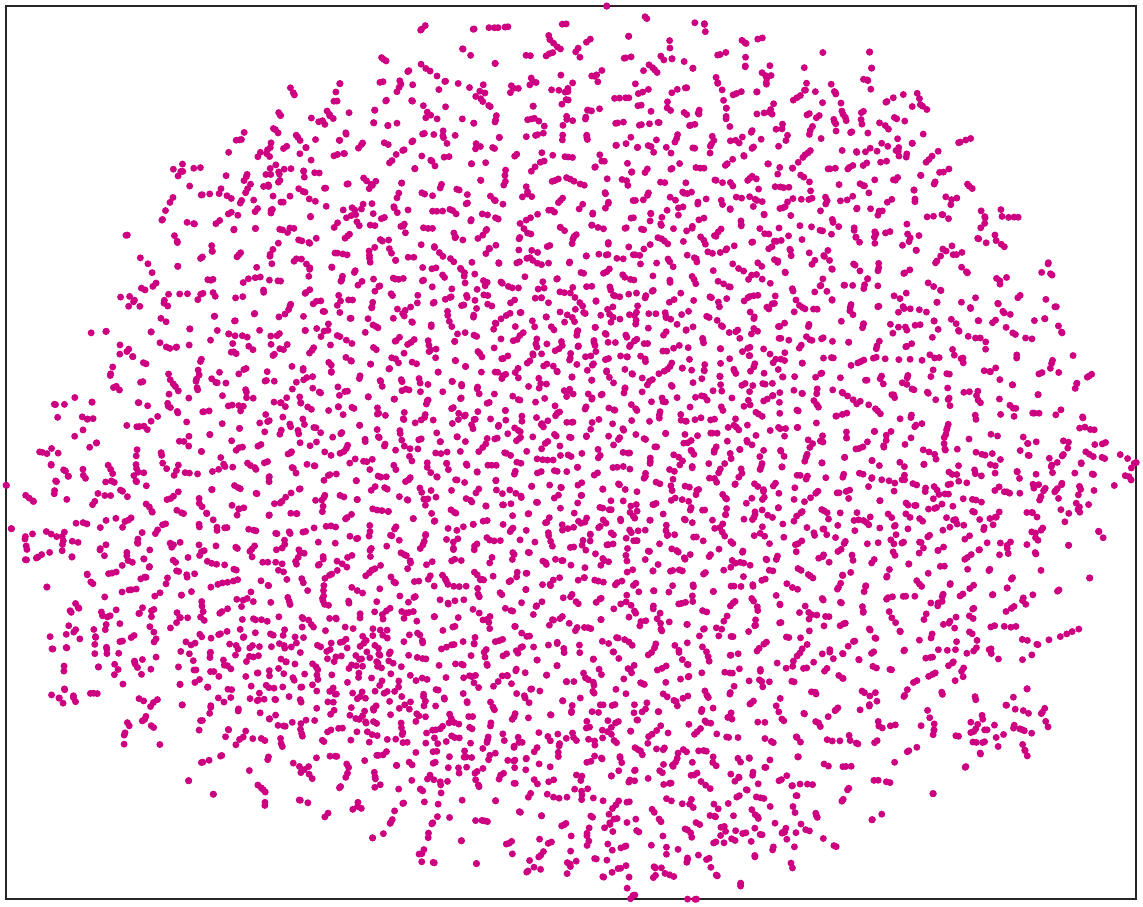}\\
   (b)  SAE (ours) 
    %
   %\arrayrulecolor{blue}\hline
   \end{tabular}
\end{center}\vspace{-0.3cm}
   \caption{Visualization of inferred codes $\bm z$ on CelebA with t-SNE. We randomly sample 5,000 faces from CelebA for illustration. The distribution of the latent codes from the variational inference (VAE) shows a standard normal one. However, the distribution of the latent codes from the spherical inference (SAE) is prone to be globally uniform while maintaining the variation of density.  }%\vspace{-0.3cm}
\label{fig:visualization1}
\end{figure*}

\clearpage
\onecolumn
\section{Reconstructed Faces with Dimension Growing}
\begin{figure*}[ht]
%\vskip 0.2in
\begin{center}
   \includegraphics[scale=0.62]{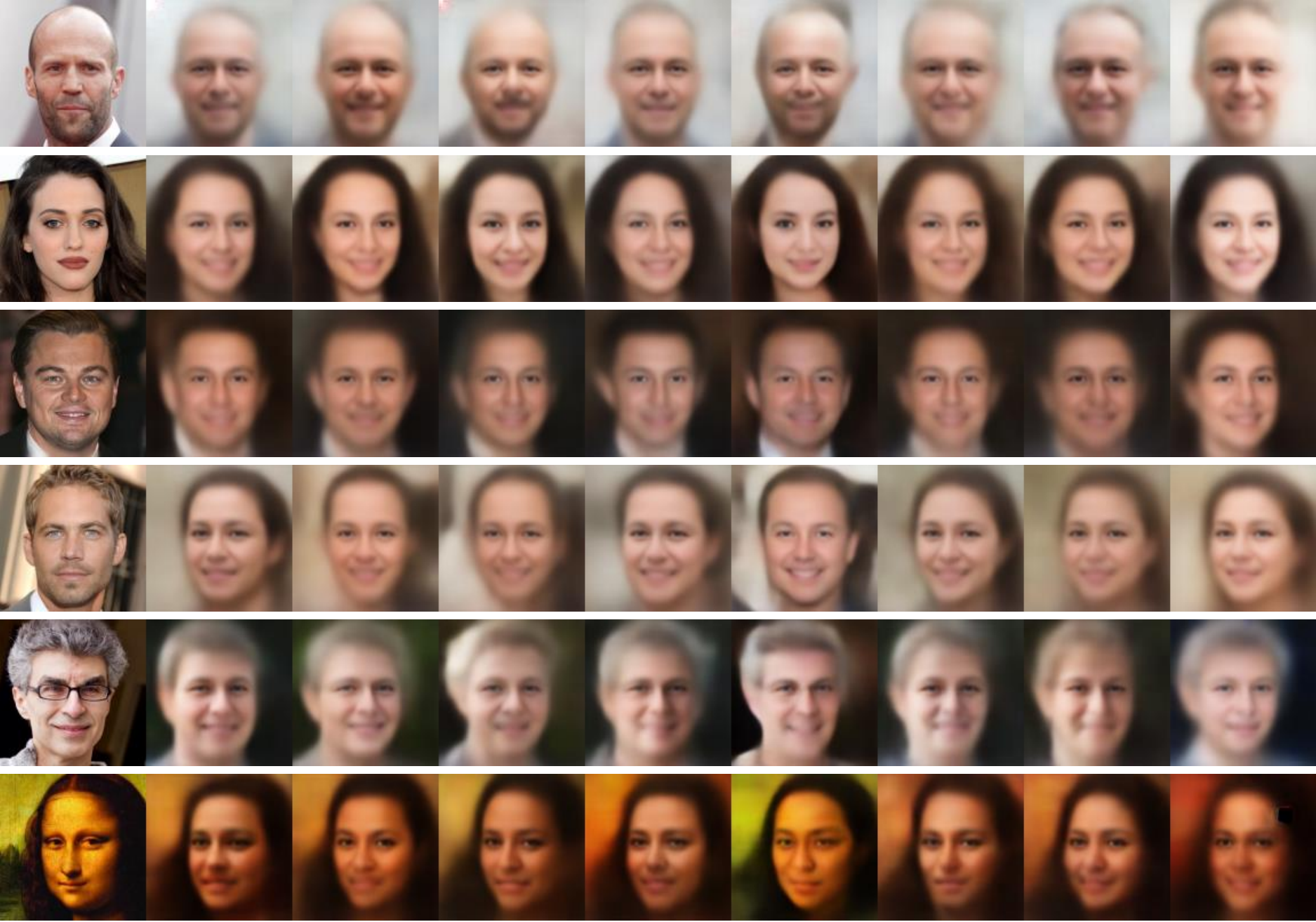}\\
   VAE \\
    \includegraphics[scale=0.62]{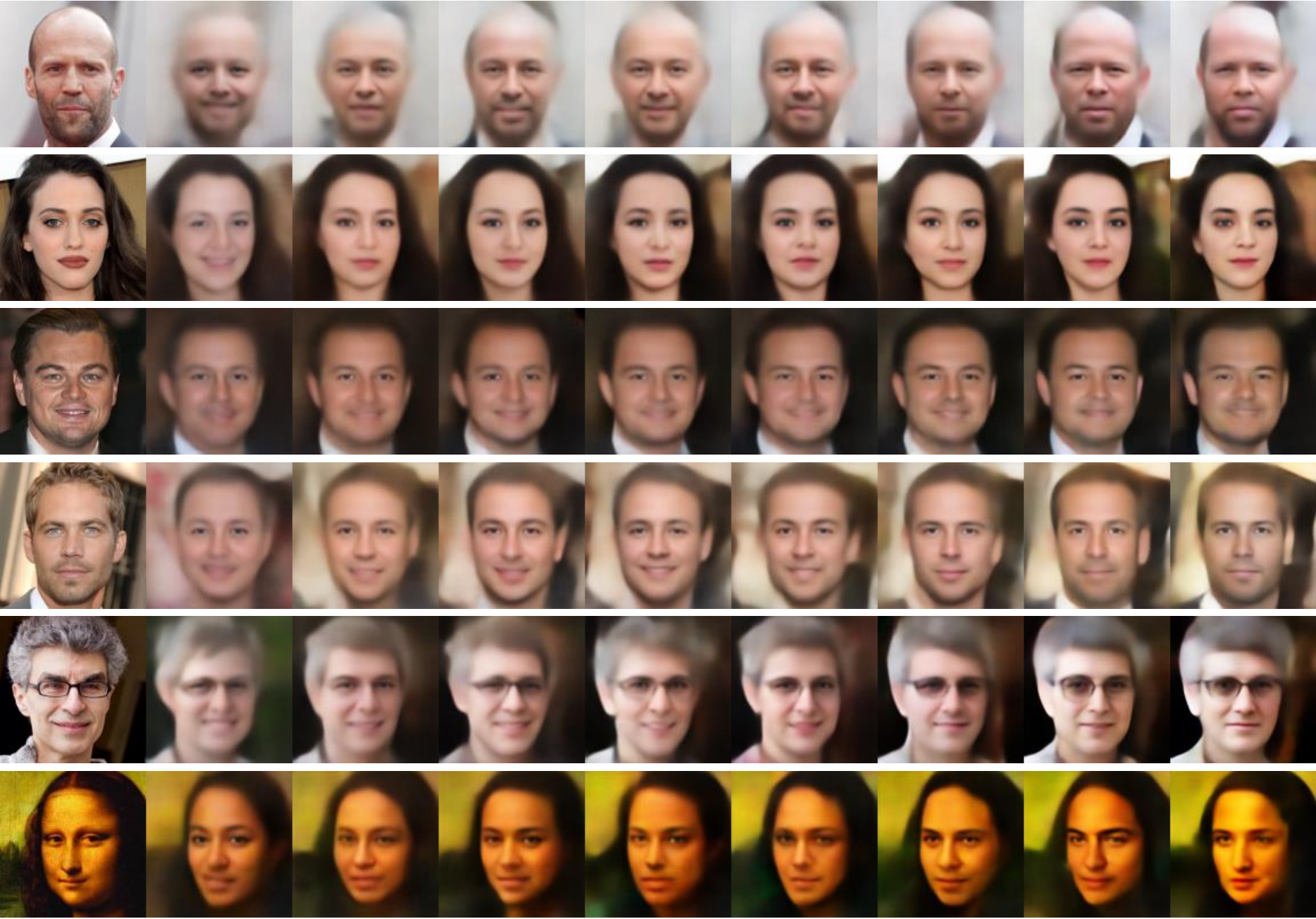}\\
    SAE\\
\end{center}\vspace{-0.3cm}
   \caption{Reconstructed faces associated with latent dimensions $2^{\{5,\dots,12\}}$. The faces in the first column are the original ones.}\vspace{-0.3cm}
\label{fig:dimension-face-sm}
\end{figure*}

\end{document}